\documentclass[conference,a4paper]{IEEEtran}
\IEEEoverridecommandlockouts
\usepackage{cite}
\usepackage{amsmath,amssymb,amsfonts}
\usepackage{algorithmic}
\usepackage{graphicx}
\usepackage{textcomp}
\usepackage{xcolor}
\usepackage{comment}
\usepackage{hyperref}
\def\BibTeX{{\rm B\kern-.05em{\sc i\kern-.025em b}\kern-.08em
    T\kern-.1667em\lower.7ex\hbox{E}\kern-.125emX}}
\begin{document}

\title{Gaze Detection and Analysis for Initiating Joint Activity in Industrial Human-Robot Collaboration*}

\author{
\IEEEauthorblockN{Pooja Prajod$^{1}$, Matteo Lavit Nicora$^{2,3}$, Marta Mondellini$^{2}$, Giovanni Tauro$^{2}$, }
\IEEEauthorblockN{Rocco Vertechy$^{3}$, Matteo Malosio$^{2}$, Elisabeth Andr\'e$^{1}$}
\IEEEauthorblockA{$^{1}$Human-Centered Artificial Intelligence, University of Augsburg, Augsburg, Germany.}
\IEEEauthorblockA{$^{2}$National Research Council of Italy, STIIMA Institute, Lecco, Italy.}
\IEEEauthorblockA{$^{3}$Industrial Engineering Department, University of Bologna, Bologna, Italy.}
\tt{pooja.prajod@uni-a.de}
\thanks{*This project has received funding from the European Union’s Horizon 2020 research and innovation programme under grant agreement No 847926.}
}

\maketitle

\begin{abstract}
Collaborative robots (cobots) are widely used in industrial applications, yet extensive research is still needed to enhance human-robot collaborations and operator experience. 
A potential approach to improve the collaboration experience involves adapting cobot behavior based on natural cues from the operator.
Inspired by the literature on human-human interactions, we conducted a wizard-of-oz study to examine whether a gaze towards the cobot can serve as a trigger for initiating joint activities in collaborative sessions. 
In this study, 37 participants engaged in an assembly task while their gaze behavior was analyzed. 
We employ a gaze-based attention recognition model to identify when the participants look at the cobot.
Our results indicate that in most cases (84.88\%), the joint activity is preceded by a gaze towards the cobot.
Furthermore, during the entire assembly cycle, the participants tend to look at the cobot around the time of the joint activity.
To the best of our knowledge, this is the first study to analyze the natural gaze behaviour of participants working on a joint activity with a robot during a collaborative assembly task.
\end{abstract}


\section{Introduction}
With the rise of the concept of Industry 4.0 and the resulting widespread adoption of cobots, the working conditions are changing rapidly~\cite{9490032}.
Therefore, there is a growing need to research the experience of operators that are now working with cobots in order to increase their well-being and reduce the risk of social isolation~\cite{nicora2021human}.

One of the crucial aspects of designing a human-robot collaborative production system is the tuning of the assigned workload since it can significantly impact the operator's well-being.
For example, a high workload is associated with distress, high blood pressure, and other indicators of low well-being~\cite{ilies2010psychological}. 
On the other hand, boredom at work leads to distress and counterproductive work behavior~\cite{van2014boredom}. 
Due to these considerations, it is important to adapt the production rhythm to the level of productivity of the operators.

In non-industrial settings, for instance in hospitals or elderly care, studies show that robotic solutions can be effective in reducing social isolation~\cite{sarabia2018assistive}.
Extending this concept to the industrial context, a cobot capable of interacting with the operator in a natural and social manner may be effective in reducing social isolation. 
To achieve such a goal, human-robot collaboration strategies should be based on everyday human-human interactions, which rely on a variety of perceptual cues~\cite{bull1985body,argyle1994gaze,hadar1983kinematics}. 
For instance, individuals instinctively direct their gaze towards their intended collaborators before initiating collaborative activities~\cite{cary1978role}.
If such behavior can be elicited during interactions with cobots, gaze direction can serve as a natural cue to communicate the intention to collaborate.

In fact, such a solution holds promise for real-time adaptation of the production rhythm to the user while, at the same time, providing social experiences akin to working with a human colleague. 
To this end, we perform an analysis of the natural gaze behaviour of participants collaborating with a cobot in an assembly task. 
A novel aspect of our study is the joint activity setup, where the human and the robot manipulate the object together.
Previous studies~\cite{huang2016anticipatory, shi2021gazeemd} have investigated gaze behaviour for industrial applications, however, the task usually involves either the human or the robot picking an object, but not lifting it together.
Our analysis supports the feasibility of using automatic attention recognition in industrial collaborative scenarios to enhance operator experience.

\begin{figure*}[thpb]
    \centering
    \includegraphics[width=0.90\textwidth]{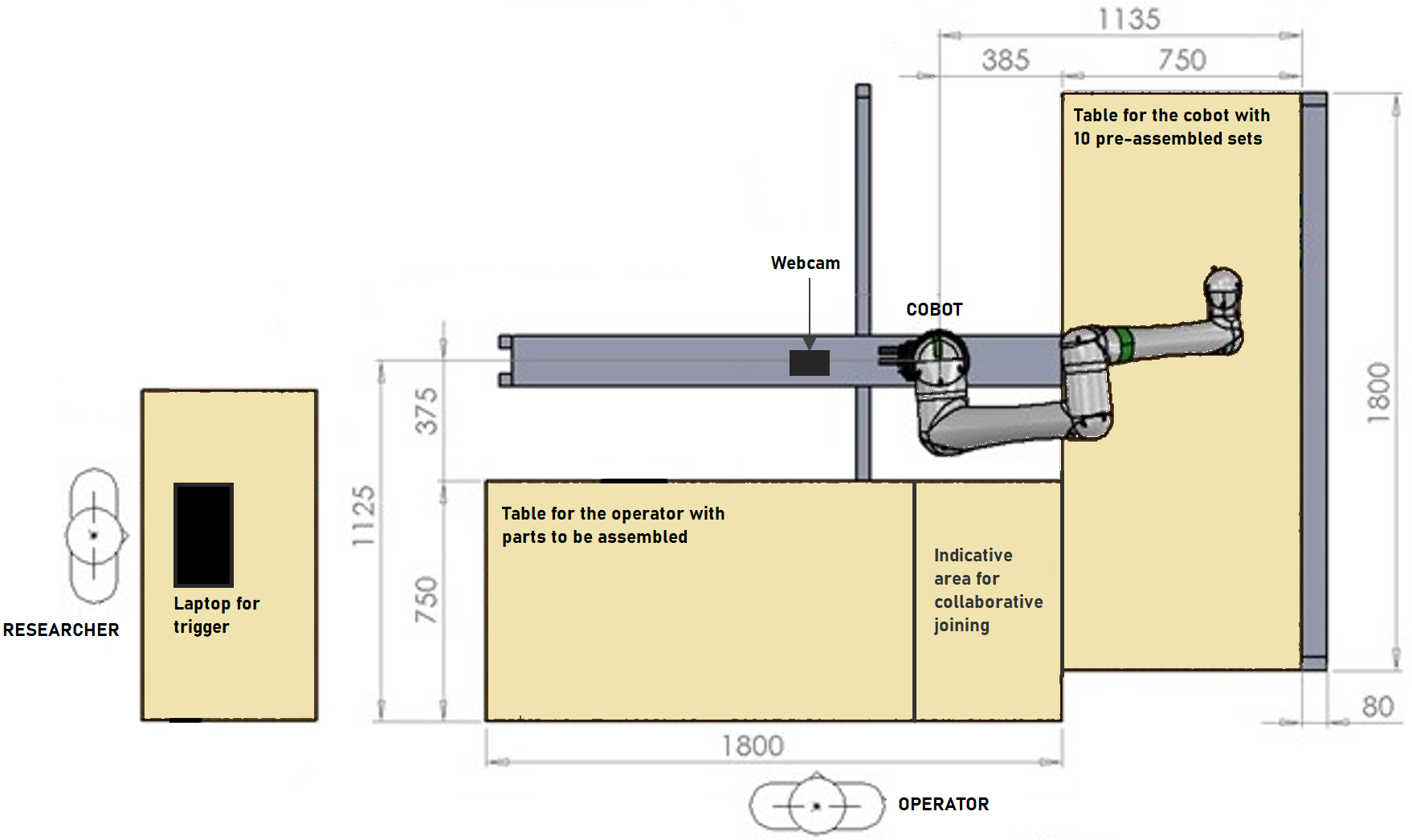}
    \caption{Schematic top-view of the experimental workcell.}
    \label{fig:workcell}
\end{figure*}

\section{Background and Related works}
\label{sec:background}

\subsection{Gaze in Human-Human Interactions}
\label{subsec:gbehavior}
Gaze is one of the communicative signals used from birth, and the number of scientific studies in this regard is really high. Gazing at a person or an object is an apparently simple act that implies at first the ability to synchronize the movements of the eyes, head, and body. 
With cognitive development, infants start to use intentional communication \cite{camaioni1992mind}, and eye contact becomes a common precursor to initiating joint attention, namely the shared focus of two individuals on an object \cite{hamilton2016gazing}. 
In this regard, Cary \cite{cary1978role} underlined that direct eye-gaze displays the willingness to interact; in particular, he watched videos of 80 students who didn't know each other, inside a waiting room. 
What emerged was that when two people started a conversation, this was almost always preceded by eye contact. 
Ferris et al. \cite{ferri2011social} conducted a series of experiments in which a subject grasped food from the table in front of him and placed it in the mouth of a person sitting on the other side. 
They found that the direct gaze of the person in front influences the performance of the gesture, proposing that the gaze makes a social request effective (to be fed) by activating a social affordance.
Innocenti et al. \cite{innocenti2012gaze} studied the impact of the gaze on a requesting gesture (i.e. grabbing an empty glass with the implicit request to fill it). 
The study demonstrated that the mere presence of a direct gaze induced a measurable effect on the subject's response in the initial phase of the sequence.
Some authors have also studied the effect of direct gaze on neural correlates. In an examination of several theories regarding the eye contact effect, Senju and Johnson \cite{senju2009eye} propose that perceived eye contact is initially detected by a subcortical route that modulates the activation of the social brain. Therefore, eye contact is closely linked to social actions not only from a behavioral point of view but from a biological point of view.

\subsection{Gaze in Robotics}
\label{subsec:grobotics}
The analysis of gaze has already been used in the past to enhance the interaction of humans and robotic systems. Often, gazing capabilities have been implemented within humanoid robots in order to expand on their social appearance~\cite{admoni2017social} and to make them more predictable in their collaborative actions~\cite{boucher2012reach}. 
However, this study focuses on the analysis of the natural gaze behavior of human participants in HRI scenarios.
Huang and Mutlu~\cite{huang2016anticipatory} designed a setup where the robot picked the pieces selected by the user.
They demonstrated that collaboration performance improves when the robot can anticipate the user's choice based on their gaze behaviour. 
Shi et al.~\cite{shi2021gazeemd} used a similar setup to demonstrate how to recognize the user's intention to pick an object solely based on their gaze behaviour.
Mehlmann et al.~\cite{mehlmann:et:al:2014} showed that a robot able to follow the user's referential gaze sped up a collaborative sorting task, reduced the number of placement attempts, and required fewer clarifications to resolve misconceptions.
Palinko et al.~\cite{palinko2016robot} studied the effectiveness of gaze information in facilitating a collaborative task.
A specific gaze sequence inspired by joint attention in human-human interaction triggered the robot's behavior. Even though the humans did not know what would activate a particular behavior of the robot, the eye-tracking capabilities of the robot helped the human complete their task. The authors also showed that eye-tracking led to more efficient human-robot interaction than head-tracking.
Saran et al.~\cite{saran2018human} trained a deep-learning model to track the user's gaze from a robot's perspective of a robot and demonstrated that it is possible to determine the user's attention to an object or the robot in real-time without dedicated eye trackers. 

In some of the existing studies, gaze behaviour serves a functional role (e.g., communicating a choice) and is often required to complete the task.
Moreover, the emphasis is typically on performance (e.g., faster completion, lower number of trials).
In our study, we explore gaze as a social cue that naturally occurs during human-robot collaboration.
In our previous work~\cite{mondellini2023behavioral}, we observed gaze behavioral patterns during a collaborative task similar to the current study task. However, these patterns were associated with waiting behavior and were not analyzed as a cue for initiating collaboration.
To the best of our knowledge, this is the first study that attempts to analyze the natural gaze behaviour of participants as a cue to initiate a joint activity in human-robot collaboration.

\section{Approach}
\label{sec:approach}


\subsection{Experimental Setup}
\begin{figure}[thpb]
    \centering
    \includegraphics[width=0.45\textwidth]{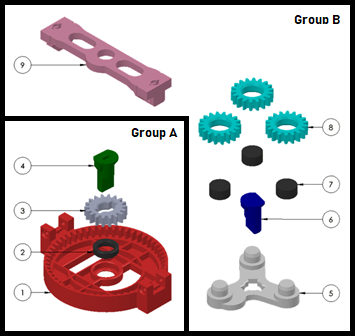}
    \caption{Preassembled components for the cobot (Group A) and components assigned to the operator (Group B).}
    \label{fig:components}
\end{figure}

To acquire the videos required for the envisioned analysis, a collaborative industrial scenario was created in a lab environment. 
The setup was made up of two distinct areas where the cobot and the operator work on their own sub-assemblies and a common area for collaborative joining to happen. With reference to Figure~\ref{fig:workcell}, an L-shaped formation was used to create said zones, plus a separate workstation for the researcher to start the system and act as Wizard of Oz during the experimental session. As shown, the wizard table is positioned on the opposite side of the cell with respect to the cobot working area so that if the operator's gaze is directed towards the wizard, this behaviour can be clearly identified and distinguished from a gaze towards their assembly table or towards the cobot itself. A 3D-printed planetary gearbox was used as a product to be assembled collaboratively by the operator and the cobot~\cite{davide_felice_redaelli_2021_5675810}. With reference to Figure~\ref{fig:components}, four components are assigned to the cobot (Group A), and the remaining five components are instead to be assembled by the operator (Group B). In order to have more freedom in the timing at which the robot is ready to bring the subassembly towards the user, the components of Group A have been pre-assembled and ten of these sets are placed on the table of the cobot, ready to be picked up as shown in Figure~\ref{fig:task}.

\subsection{Demographics and Assembly Task}
\begin{figure*}[thpb]
    \centering
    \includegraphics[width=0.99\textwidth]{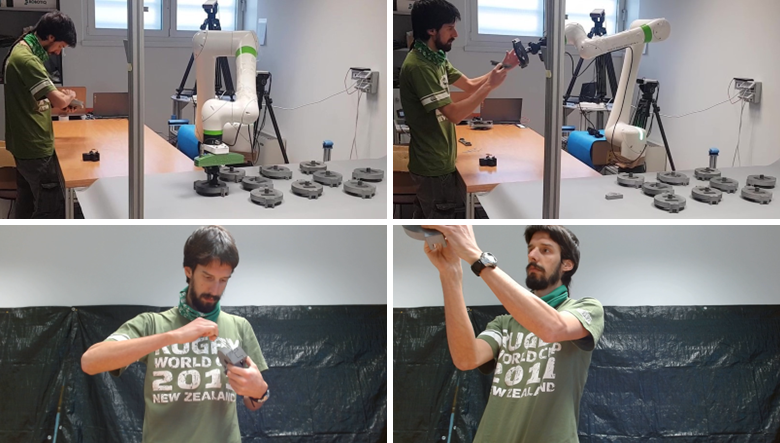}
    \caption{The images show the joint activity between the cobot and the participant and a no collaboration instance from two different viewpoints. On the top is an overview of the setup from the side. On the bottom is the front camera recordings that were used in the analysis.}
    \label{fig:task}
\end{figure*}

A total of 37 adult volunteers took part in the experiment (29 males) ranging from 18 to 48 years old. Prior to engaging in the assembly task, each participant received appropriate training. The experiment session duration of 15 minutes was carefully chosen to ensure an adequate number of assembly cycles (approximately 15 to 20 complete products) for each participant, enabling a comprehensive analysis of their recurring gaze behavior. Looking at Figure~\ref{fig:components}, each participant had to assemble Group B while the robot would hover with the detection camera over the pre-assembled Group A. As the volunteer's task got close to completion, the wizard pressed a button on the laptop to trigger the robot to pick up the subassembly and bring it in front of the user at a convenient angle for the final joining, as shown in Figure~\ref{fig:task}. This iterative process continued throughout the 15-minutes experimental session, regardless of the number of completed gearboxes. To ensure a smooth workflow, ten pre-assembled sub-assemblies were initially placed on the cobot's table. The researcher restocked the sub-assemblies as necessary. Importantly, participants were unaware of the trigger given by the researcher to prevent any potential biases in their behavior during the interaction with the cobot. Also, the participants were informed of being filmed for ethical reasons but the aim of studying their gaze behaviour was revealed only at the end of the session, again to avoid any possible bias.

\subsection{Ethical Approval}
The study has been conducted according to the guidelines of the Declaration of Helsinki and approved by the Ethics Committee of I.R.C.C.S. Eugenio Medea (protocol code N. 19/20-CE of 20 April 2020).

\subsection{Tools}
\label{sec:tools}
\subsubsection{Attention Recognition Model}
While piloting the setup, we identified three main areas of interest in the environment: the cobot, the  table (looking at the table while assembling), and anywhere else (looking at the clock, window, etc.).
Consequently, we trained a deep learning model that classifies the gaze direction into these three areas.
First, we trained a convolutional neural network (VGG16 architecture) using ETH-XGaze dataset~\cite{xgaze} for gaze estimation.
Then, we fine-tuned the last few layers of this model to map the gaze to one of the areas of interest.
This model achieves an accuracy $=$ 94.3\% and an f1-score $=$ 94\%.
The details about the training procedure and validation of this model can be found in~\cite{prajod2023petra}.
We also implemented a Social Signal Interpretation (SSI)~\cite{WagnerSSI} pipeline capable of inferring information related to what the subject is paying attention to for each frame of an input video stream.
 
\subsubsection{NOVA Annotation}
The participants perform two primary activities: assembling their own sub-assembly and joining the sub-assemblies along with the cobot (joint activity). 
In this study, we focus on the gaze behaviour of the participants, especially the few seconds leading up to the joint activity.
For each assembly cycle, we annotate the frame where the cobot arrives for the joint activity.
We use the NOVA tool~\cite{baur2013nova} to annotate these frames. 
NOVA also allows us to visualize the predictions from the attention recognition model as a stream.

\section{Analysis}
\label{sec:analysis}
In human-human interactions, gaze-based social cues facilitate collaboration.
For example, the interaction is often initiated by looking at the other person.
However, it is not known whether humans naturally exhibit similar gaze behaviour when collaborating with an industrial cobot.
To this end, we analyze the gaze behaviours of the participants working with a cobot on a collaborative assembly task.
Specifically, we investigate if the participants gaze towards the cobot to initiate the collaborative joining of sub-assemblies.
We note that the wizard controlled the cobot using the information about the completion of the sub-assembly and not their gazes.
Hence, the participants were not required to exhibit any gaze pattern to complete the task.
Moreover, they did not know what event triggers the cobot for joint activity.
This setup allows us to analyze the natural gaze behaviours of the participants collaborating with a cobot, especially how they attempt to initiate the joint activity. 

\begin{figure*}[htpb]
    \centering
    \includegraphics[width=0.99\textwidth]{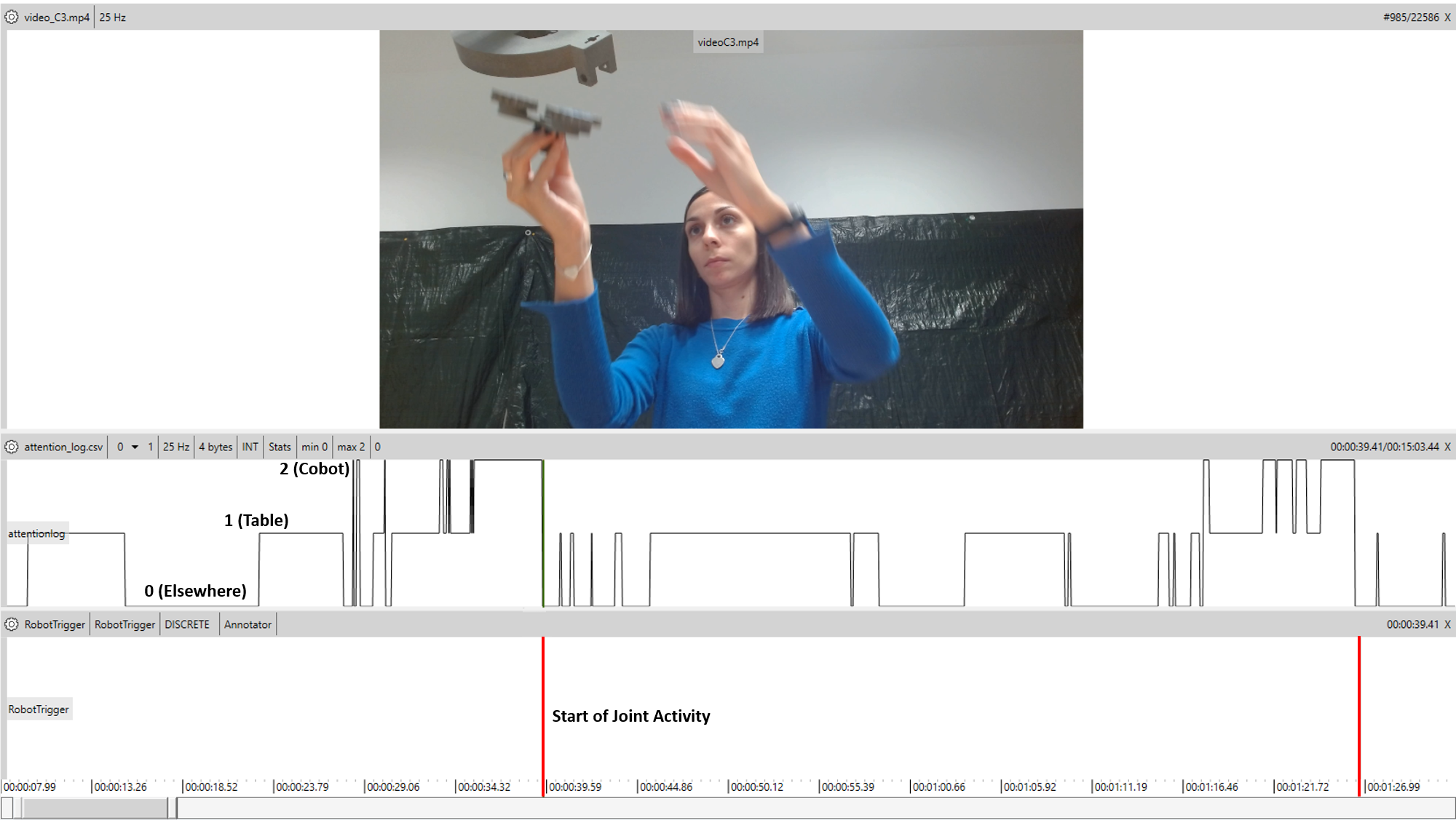}
    \caption{A snapshot from the NOVA tool showing the predictions from the attention recognition model (top track), and the annotated joint activity start points (bottom track, red lines).}
    \label{fig:nova_vis}
\end{figure*}

We use an attention recognition model (see Section~\ref{sec:tools}) to classify the gaze into three classes (0 - random, 1 - table, 2 - cobot).
This model saves the annotation efforts involved in manually labeling the entire video.
We use the NOVA tool (see Section~\ref{sec:tools}) to annotate the frame where the cobot arrives for the collaborative joining of the sub-assemblies.
This point is considered the start of the joint activity in each assembly cycle.
In addition, as shown in Figure~\ref{fig:nova_vis}, we use NOVA to visualize the predictions from the attention recognition model along with the joint activity start points.
The bottom track shows the annotated starting points of the joint activity.
The values in the top track can be 0, 1, or 2 depending on the predicted class. 
We specifically focus on the instances where the predicted class is 2, i.e., the gaze is predicted towards the cobot.
A promising trend is observed as spikes (class = 2) in the top track in the few seconds leading up to the joint activity.
This pattern indicates that the participant is looking at the cobot plausibly to initiate the joint activity.

We analyze this gaze pattern for each participant in two steps.
First, we calculate the gazes to the cobot within 15 seconds prior to the joint activity.
We chose 15 seconds because, after the trigger, the cobot takes 10-12 seconds to move over the part, grab it, pick it up and bring it to the collaborative joining position (3 seconds).
This step helps us determine how often the joint activity is preceded by gazing towards the cobot, and therefore a cue to initiate the activity.
Second, we calculate the gazes to the cobot that are outside the above-mentioned 15 seconds and also outside the joint activity itself.
This step allows us to make sure that the gaze pattern is prominent around the time of the joint activity, and not a frequent behaviour irrespective of the activity.

Before calculating the gazes, we smooth the predictions from the attention recognition model using a three-point moving window.
We use a peak detection algorithm to find the points where the gaze is towards the cobot.
We only consider the peaks which span for at least five frames, i.e., the participant looks at the cobot for at least five consecutive frames.
Using these peak points and the annotated starting points of the joint activity, we calculate the percentage of gaze-preceded joint activities (\textit{pGazeJoint}) and the percentage of unexpected gazes to the cobot (\textit{pUnexpectedGaze}). 
Joint activity is deemed gaze-preceded if the participant looks at the cobot at least once during the 15 seconds prior to the start of the joint activity.
So, \textit{pGazeJoint} (expressed in percentage) is the number of gaze-preceded joint activities out of the total joint activities in the session.
We expect the participants to look at the cobot for initiating the joint activity and for the duration of the activity (typically lasts for 20-25 seconds).
Any gaze towards the cobot that occurs outside this duration is considered unexpected.
We calculate \textit{pUnexpectedGaze} (expressed in percentage) as the ratio of unexpected gazes towards the cobot to the total number of gazes towards the cobot.

\begin{figure}[htpb]
    \centering
    \includegraphics[width=0.37\textwidth]{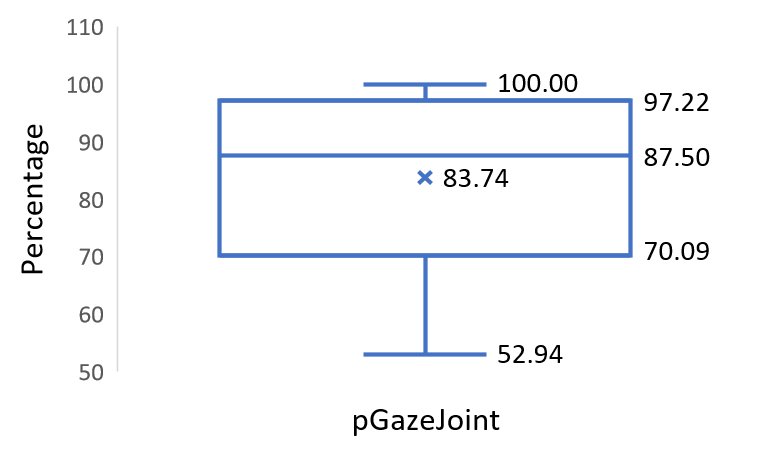}
    \caption{Box-plot visualization of \textit{pGazeJoint} values computed from 37 participants}
    \label{fig:pGC_boxplot}
\end{figure}

\begin{figure}[htpb]
    \centering
    \includegraphics[width=0.37\textwidth]{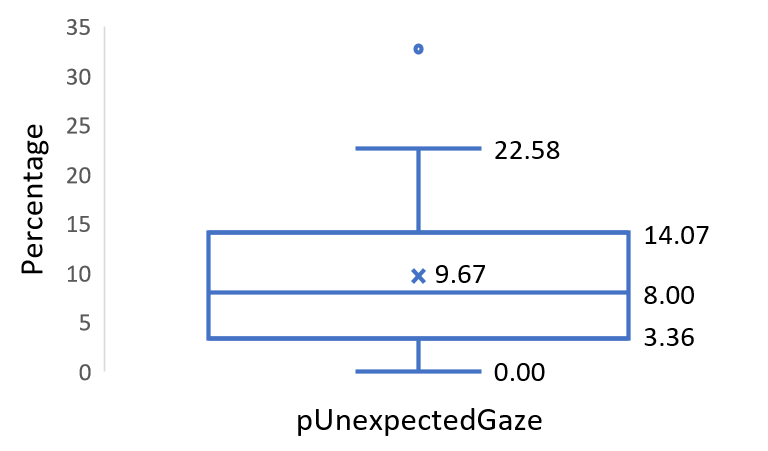}
    \caption{Box-plot visualization of \textit{pUnexpectedGaze} values computed from 37 participants}
    \label{fig:pUG_boxplot}
\end{figure}

Figures~\ref{fig:pGC_boxplot} and~\ref{fig:pUG_boxplot} visualizes the \textit{pGazeJoint} and \textit{pUnexpectedGaze} values from 37 participants as box-plots.
The mean \textit{pGazeJoint} value is 83.74, i.e., on average, 83.74\% of all collaborative joining instances were preceded by a gaze towards the cobot.
Similarly, the mean \textit{pUnexpectedGaze} is 9.67\%, which implies that only very few gazes at the cobot were outside the expected time frame.
In other words, looking at the cobot occurs predominantly around the time of the collaborative joining activity. 
These results indicate that people use gaze as a social cue to initiate joint activity even when interacting with a cobot.

\section{Discussion}
\label{sec:discussion}

Our results show that people tend to look at the cobot when they are ready to work jointly on a task (represented by high \textit{pGazeJoint}), a behaviour prevalent in human-human interaction.
This behaviour can be seen as a social cue to initiate a joint activity, thereby promoting a more natural and intuitive human-robot collaboration.

Additionally, our results indicate that gaze directed at the cobot typically occurs during the collaborative joining activity or shortly before the start of the joint activity, represented by low \textit{pUnexpectedGaze}.
We observed that longer joining times were one of the factors contributing to unexpected gazes towards the cobot. 
Specifically, during certain assembly cycles, participants took more time to align the sub-assemblies, resulting in a collaborative joining process that exceeded the estimated duration.
Furthermore, errors or delays in the cobot's performance were also responsible for unexpected gazes. 
For instance, in some cases, the robot did not initiate the subsequent assembly cycle immediately after completing the previous one due to unexpected software behaviors. 
Consequently, a few seconds of unforeseen delay preceded the next series of robot movements, capturing the participants' attention and prompting them to look towards the robot to comprehend the situation.

We received some insightful comments (translated from Italian) from the participants after the experiments.
One of the participants (Participant 3) said: "I noticed that the robot was synchronized with me and I thought it might be because of the camera, so I tried looking at it to see what would happen".
Another participant (Participant 36) said: "In some cases, I was surprised by how slow the robot was, so I tried looking at it in the hope to make it faster".
These participants inferred that their gaze influenced the cobot's behaviour; whereas in reality, it solely relied on the wizard's judgment of whether the participant completed their sub-assembly.
These comments further reinforce the idea of using gaze to facilitate more natural human-robot collaboration.

Participant 14 provided an interesting suggestion about adding eyes to the cobot to make it expressive.
Although this suggestion relates to anthropomorphism and is beyond the scope of this work, it highlights a possible direction to make human-robot collaboration more natural.

\section{Conclusions and Future works}
\label{sec:conclusions}

The collaboration experience of cobot workers can be improved by incorporating elements from human-human interactions.
In this work, we investigate if people gaze towards the cobot as a cue to initiate a joint activity.
Although this behaviour is common in human-human interactions, it is not known if such behaviour occurs during human-robot collaborations.
To this end, we study the gaze behaviours of 37 participants collaborating with a cobot in an industry-like assembly task.
We use a wizard-of-oz setup to trigger the collaborative joining activity.
Using a gaze-based attention recognition model, we identify the instances where the participant looks at the cobot.
Our analysis shows that 84.88\% of the joint activities were preceded by a gaze towards the cobot.
We also find that, in the entire assembly cycle, the participants tend to look at the cobot around the time of the joint activity.
Our results indicate that the gaze-based initiation cue indeed extends to human-robot collaboration.
Hence, it is feasible to use natural cues from the worker for adapting the cobot behaviour to improve the worker's experience.

In the future, we will study if the gaze-based initiation cue is valid in longer collaboration sessions and in a real-life setting (e.g. actual industrial workcell).
For instance, the participants may start expecting the cobot to know the appropriate time for joint activity, even without any cues from the participant.
We also aim to implement real-time adaptation of the cobot behaviour (e.g., tuning the cobot production rate to the operator's rhythm) based on the automatic detection of the gaze initiation cue.
Subsequently, we will explore the benefits of the adaptive behaviour of the cobot in terms of the well-being and experience of the operators.

\bibliographystyle{IEEEtran}
\bibliography{IEEEabrv,bibfile}

\end{document}